\documentclass[10pt,twocolumn,letterpaper]{article}
\usepackage{booktabs}
\usepackage{cvpr}
\usepackage{cite}
\usepackage{times}
\usepackage{epsfig}
\usepackage{diagbox}
\usepackage{graphicx}
\usepackage{amsmath}
\usepackage{caption}
\usepackage{pifont}
\usepackage{float}
\usepackage{amssymb}
\usepackage{enumitem}
\usepackage{multirow}
\usepackage{bm}
\usepackage{array}

\usepackage[breaklinks=true,bookmarks=false]{hyperref}

\cvprfinalcopy % *** Uncomment this line for the final submission

\def\cvprPaperID{****} % *** Enter the CVPR Paper ID here

\begin{document}

%%%%%%%%% TITLE
\title{A Universal Railway Obstacle Detection System based on Semi-supervised Segmentation And Optical Flow}

\author{Qiushi Guo\\
CSRD\\
{\tt\small  
 guoqiushi@csrd.cn}}
% \thanks{$\dag$ indicates corresponding author}

\maketitle
%\thispagestyle{empty}
%%%%%%%%% ABSTRACT
\begin{abstract}
Detecting obstacles in railway scenarios is both crucial and challenging due to the wide range of obstacle categories and varying ambient conditions such as weather and light. Given the impossibility of encompassing all obstacle categories during the training stage, we address this out-of-distribution (OOD) issue with a semi-supervised segmentation approach guided by optical flow clues. We reformulate the task as a binary segmentation problem instead of the traditional object detection approach. To mitigate data shortages, we generate highly realistic synthetic images using Segment Anything (SAM) and YOLO, eliminating the need for manual annotation to produce abundant pixel-level annotations. Additionally, we leverage optical flow as prior knowledge to train the model effectively. Several experiments are conducted, demonstrating the feasibility and effectiveness of our approach.
\end{abstract}
\section{Introduction}
% With the advancement of high-speed trains, the security risks associated with railway systems have become a major public concern. Obstacle detection remains a significant challenge in railway security. Creating a reliable and scalable obstacle detection system can assist drivers and dispatchers in making proactive decisions to prevent accidents. Deep learning has been extensively applied in various security-related fields, such as mobile payments, disaster detection\cite{flood2019detecting}, and fraud detection\cite{guo2023enhancing}, showing great potential to improve railway safety through advanced obstacle detection technologies.

With the rapid advancement of high-speed trains, ensuring the security of railway systems has emerged as a critical public concern. One of the primary challenges is obstacle detection, which plays a crucial role in railway safety. Developing a reliable and scalable obstacle detection system can empower train operators and dispatchers to take preemptive actions and mitigate potential accidents.

Deep learning techniques have been widely adopted across various security domains, including mobile payments\cite{cai2022enable}, disaster detection \cite{flood2019detecting}, and fraud detection \cite{guo2023enhancing}. This technology exhibits substantial promise in enhancing railway safety through sophisticated obstacle detection capabilities. Significant efforts have recently been devoted to addressing obstacle detection using deep learning methods. Although these approaches have achieved some success, they also exhibit notable disadvantages:
\begin{figure}[htb]
    \centering
    \includegraphics[width=0.5\textwidth]{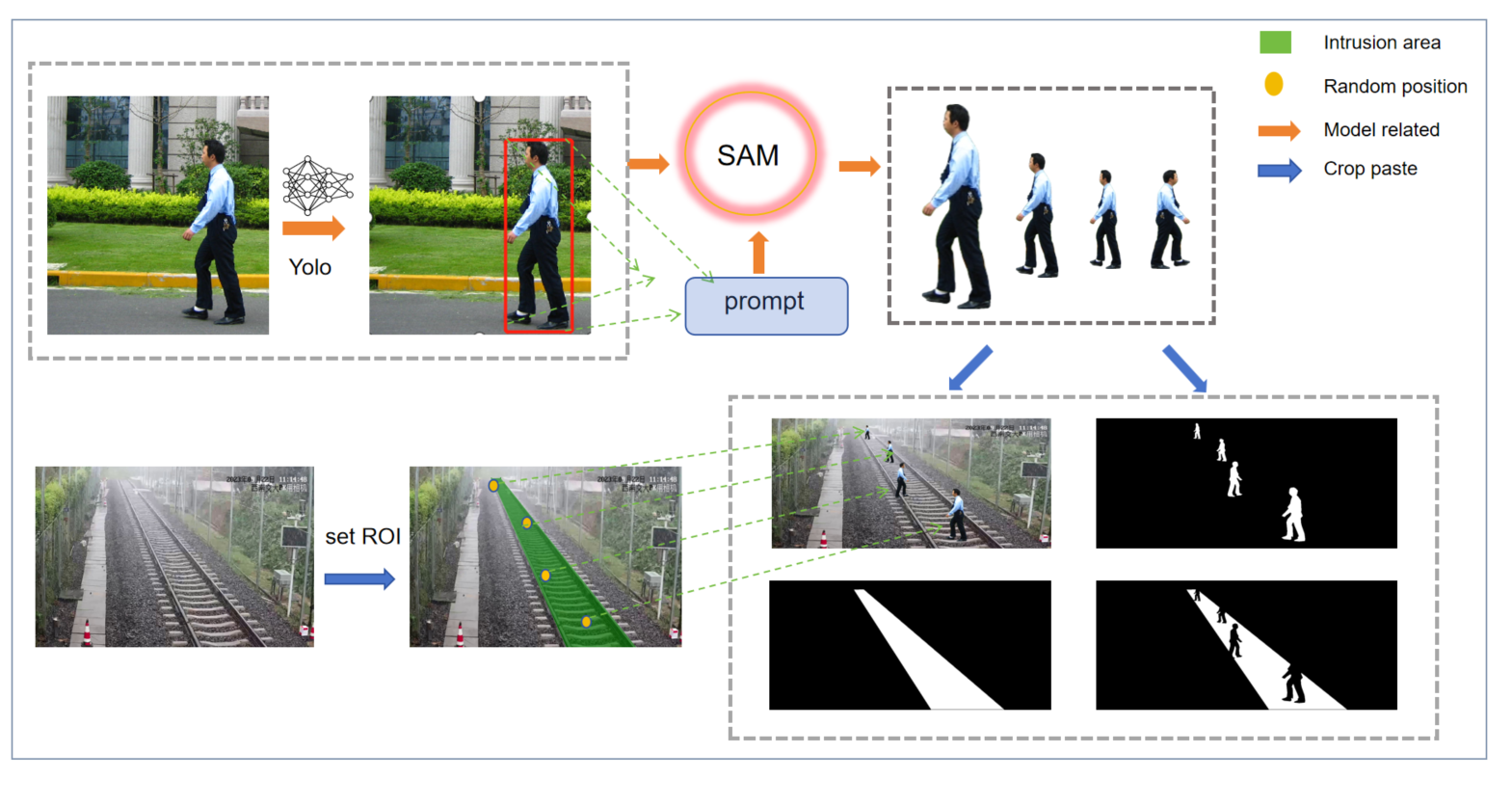}
    \caption{The pipeline for synthetic data generation, utilizing SAM and YOLO to extract target objects from a gallery and superimpose them onto a base image (specifically, a railway image). Notably, this process does not necessitate annotations.}
    \label{fig:pseudo}
\end{figure}
\begin{itemize}
    \item Fragility to complex ambient conditions
    \item Requirement for extensive manual annotations
    \item Difficulty in extending to different scenarios
\end{itemize}
Designing an extendable, annotation-free model with strong generalization ability remains a significant challenge in both industry and academia.

To address the aforementioned issues, we propose a semi-supervised approach guided by optical flow. To mitigate the data shortage problem, we employ SAM \cite{kirillov2023segment} and YOLO \cite{redmon2016you} to generate highly realistic pseudo-images for training. Instead of manually collecting and annotating images pixel by pixel, we prepare two image sets: base images (fewer than 100 background images with only railway areas annotated) and object images. The object images include categories such as pedestrians, animals, and textures. Using SAM and YOLO, we obtain masks for the intended objects in these images. These objects are then pasted onto the base images according to the masks. The entire process are  illustrated as \textbf{Fig.} \ref{fig:pseudo} This process simultaneously generates image and mask pairs without manual effort.

% To handle challenging weather scenarios, we implement two strategies. First, we collect base images under various weather conditions, including rainy, foggy, and normal (sunny) conditions. Second, we use optical flow to locate position information as prior knowledge. To obtain optical flow predictions, we generate pseudo sequences of obstacles. Specifically, we create a pseudo frame at point $P_{i}(x,y)$ and generate a new one at $P_{i+1}(x+\delta, y+\delta)$ with the same pasted object. We conduct several experiments which show that our approach demonstrate satisfied results across different scenarios.

To address the challenges posed by varying weather conditions, we implement two complementary strategies. Firstly, we compile a dataset of base images captured under diverse weather conditions, including rainy, foggy, and clear (sunny) environments. Secondly, we utilize optical flow to provide positional information as prior knowledge. For optical flow predictions, we generate pseudo sequences of obstacles. This involves creating an initial pseudo frame at point $P_{i}(x,y)$ and subsequently generating a new frame at $P_{i+1}(x+\delta, y+\delta)$ with the same object superimposed. Experimental results indicate that our approach yields satisfactory performance across different weather scenarios.

Our contributions are summarized as follows:
\begin{itemize}[topsep=0pt, itemsep=1ex, parsep=2pt]
    \item We reformulate the obstacle detection task as a binary segmentation problem, distinguishing between railway areas and non-railway areas.
    \item We introduce a simple yet effective data generation mechanism to synthesize realistic images using SAM and YOLO.
    \item Optical flow is leveraged to generate prior knowledge that guides the segmentation network.
\end{itemize}

\begin{figure*}[t]
    \centering
    \includegraphics[width=0.9\textwidth]{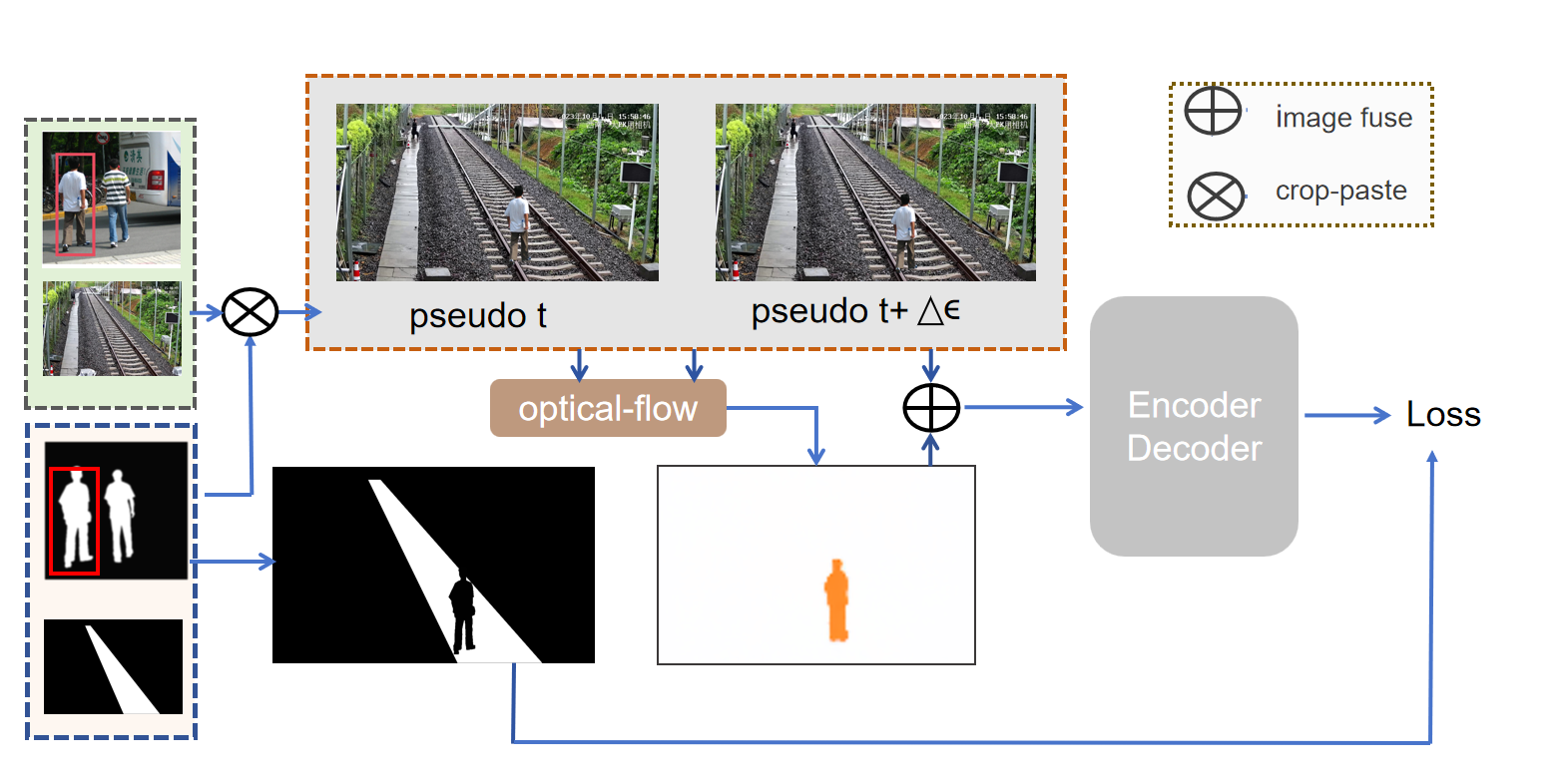}
    \caption{Pipeline of our proposed method.}
    \label{fig:pipeline}
\end{figure*}

\section{Related work}
\subsection{Obstacle Detection in Railyway}
Matthias Brucker \textit{et al}.\cite{brucker2023local}propose a a shallow netwrok to learn railway segmentation from normal railway images. They explore the controlled inclusion of global information by learning to hallucinate obstacle-free images. Zhang Qiang \textit{et al}.\cite{zhang2023automatic}. conbine segmentation model with the LiDAR in their obstacle detection system; Amine Boussik \textit{et al}.\cite{boussik2021railway} propose an unsupervised models based on a large set of generated convolutional auto-encoder models to detect obstacles on railway’s track level.
\subsection{Segmentation with Optical flow}
Laura \textit{et al}. \cite{sevilla2016optical}. demonstrate the effectiveness of jointly optimizing optical flow and video segmentation using an iterative scheme; Volodymyr \textit{et al}\cite{fedynyak2024devos}. present an architecture for
Video Object Segmentation that combines memory-based
matching with motion-guided propagation resulting in stable long-term modeling and strong temporal consistency.
\section{Method}
% Assumes that a set of base images $B$ and a set of target images $T$ are given. Our tasks is to detect all potential obstacles in certain areas $\eta$. Unlike traditional detection approach, which attempts to detect categories of each obstacle, we reformulate the problem as a binary segmentation task. Rather than detecting all potential obstacles(which is impossible), we focus on the railway area, which is consistent during all time compared to obstacles. To simulate the these scenarios, we generate high realistic pseudo images in copy-paste manner. Besides,to deal with the extreme weather conditions which is hard to segment 
% objects, optical-flow is introduce to provide prior information to guide the segmentation model. Pseudo images $I_{t}$ and $I_{t}$ are generated with a tiny shift $\delta$ of the target object to simulate the movement of target. The out of optical-flow model will be stacked along with pseudo images as input to make predictions. In this section, we will discuss the details of the entire process.
% The pipeline of our approach are demonstrated as \textbf{Fig.}\ref{fig:pipeline}. Given a set of base images $B$ and target images $T$, our task is to detect potential obstacles within specific areas $\eta$. Unlike traditional detection methods that aim to categorically detect each obstacle, we reformulate the problem as a binary segmentation task. Rather than attempting to detect all potential obstacles (which is impractical), our focus is on segmenting the railway area, which remains consistent over time compared to obstacles. 
The pipeline of our approach is illustrated in \textbf{Fig.}\ref{fig:pipeline}.
Given a set of base images $B$ and target images $T$, our objective is to identify potential obstacles within specific regions $\eta$. Unlike traditional detection methods that categorically detect each obstacle, we reformulate the problem as a binary segmentation task. Instead of attempting to detect all potential obstacles, which is impractical, our emphasis is on segmenting the railway area, a region that remains consistent over time compared to obstacles.

To simulate these scenarios effectively, we generate highly realistic pseudo-images using a copy-paste approach. Additionally, to address challenges posed by extreme weather conditions, which can obscure object segmentation, we introduce optical flow to provide prior information guiding the segmentation model. Pseudo images $I_{t}$ and $I_{t+\delta}$ are generated by applying a small shift $\delta$ to the target object, simulating its movement. The output of the optical flow model is incorporated along with pseudo images as input to facilitate accurate predictions. 
This section will delve into the detailed methodology employed throughout this process.

% \begin{figure*}[htb]
%     \centering
%     \includegraphics[width=0.9\textwidth]{figures/pipeline.png}
%     \caption{pseudo data generation}
%     \label{fig:pseudo}
% \end{figure*}

\subsection{Data Acquisition}
% \noindent\textbf{Base Image} are all collected in our experiment center in ChengDu, which contains a railway with over 60 meters and fog/rain simulators. To rich the diversity of our dataset, we collect images under various weather conditions, namely rainy, foggy and sunny. Since the camera is fixed,
% only one mask is needed to annotate. Beside, there are no possible obstacles in railway in base images. The obstacles can only be generated in copy-paste manner.

\begin{figure}[H]
    \centering
    \includegraphics[width=0.5\textwidth]{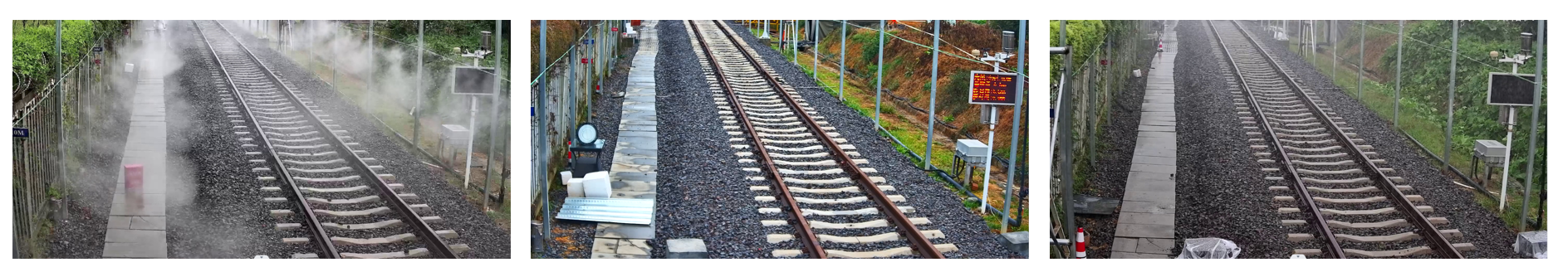}
    \caption{Sample base images depicting various weather conditions. From left to right, the images illustrate scenes captured under foggy, normal, and rainy conditions. }
    \label{weathers}
\end{figure}

\noindent\textbf{Base Images} are used in our experiments are gathered at our facility in Chengdu, which features a railway spanning over 60 meters and includes simulators for fog and rain conditions. To ensure diversity in our dataset, we capture images under different weather scenarios, specifically rainy, foggy, and sunny conditions \textbf{Fig.} \ref{weathers}. Due to the fixed position of the camera, only one mask is required for annotation purposes. Importantly, the railway areas in the base images are devoid of any potential obstacles. Any obstacles present are generated using a copy-paste method.
% \begin{figure}[t]
%     \centering
%     \includegraphics[width=0.5\textwidth]{figures/weather.png}
%     \caption{sample base images of different weathers. From left to right: Foggy, normal and rainy.}
%     \label{weathers}
% \end{figure}

% \noindent\textbf{Object Image} contain three categories: PennFudanPed, Obj365(part)\cite{shao2019objects365} and DTD\cite{cimpoi2014describing}. To make the whole process can be applied automatically, we assume no masks are provided. We select categories which may occur in our scenario with a high probability, animals like deer,horse,cow, vehicles like truck, cart, and so on. 
\noindent\textbf{Object Image} dataset comprises three categories: PennFudanPed, Obj365 (part) \cite{shao2019objects365}, and DTD \cite{cimpoi2014describing}. To facilitate fully automated application of our methodology, we proceed under the assumption that no masks are initially available. We focus on selecting categories likely to occur in our scenario, such as animals (e.g., deer, horse, cow) and vehicles (e.g., truck, cart). This ensures our approach is tailored to handle relevant objects effectively.
\begin{figure}[H]
    \centering
    \includegraphics[width=0.5\textwidth]{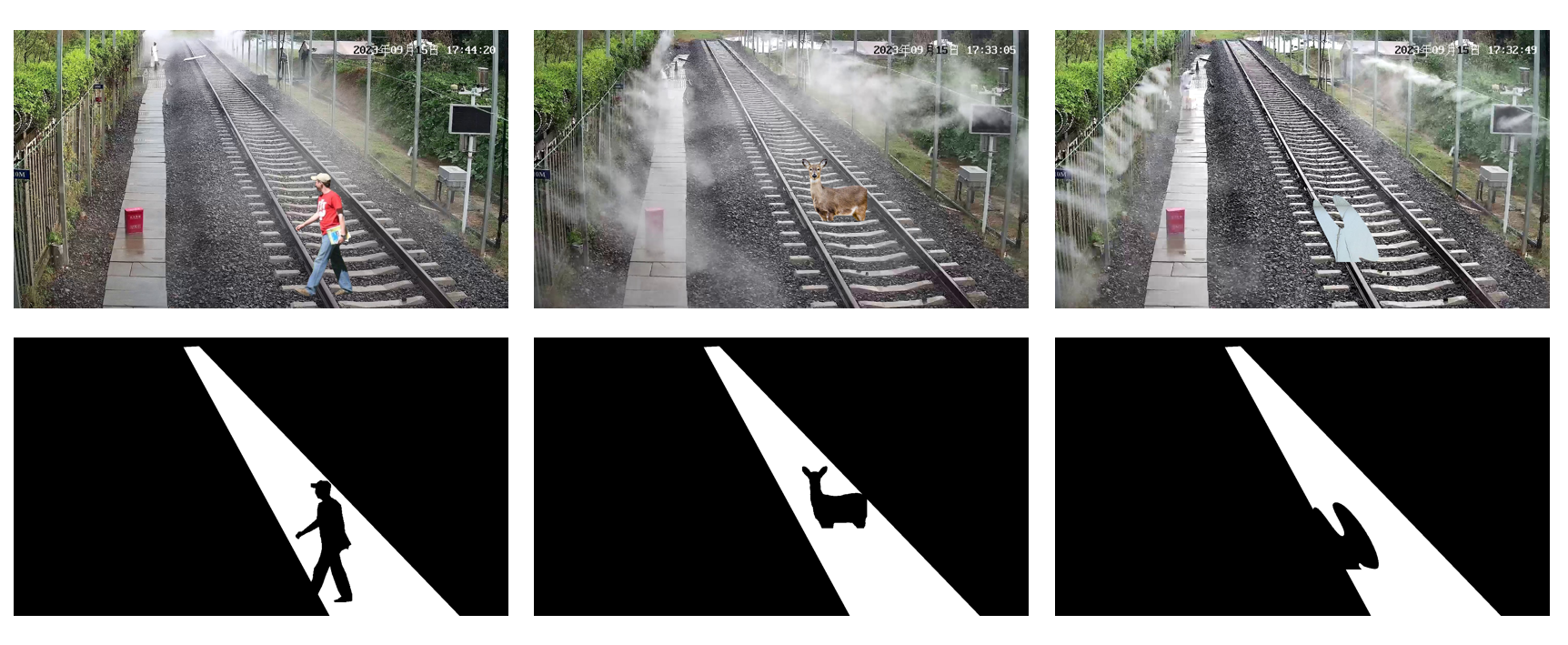}
    \caption{Sample of generated image-mask pairs. From left to right: pedestrian, animal and texture}
    \label{sample}
\end{figure}
% \begin{figure}[t]
%     \centering
%     \includegraphics[width=0.5\textwidth]{figures/sample.png}
%     \caption{sample of generated image pairs. From left to right: pedestrian, animal and texture}
%     \label{sample}
% \end{figure}
% \subsection{Copy-Paste} The entire process can be divided into follow steps: feeding the object images into the YOLO model, a list of bounding boxes of detected targets are returned. The bounding box can be leveraged as the prompt for the SAM to obtain the segmentation mask. The pixels belong to the objects will be pasted onto the base images guided by the segmentation mask. Let us dive into the details of the whole process. Only objects belong to the object-365 class list can be detected. We pre-define a target category list to fit our scenario. In each object image, only one object is selected randomly as the target object. In SAM stage, not every output mask is perfect as our imagined. But it still contributes to our purpose. Our aim is to segment railway area not obstacles. To deal with the OOD issue, we generate polygons with random shape, render it with texture images in DTD. Besides, to further imporve the robustness of our model, we rescale the objects to enrich the content of the image. 
The entire process can be delineated into sequential steps: Initially, object images are fed into the YOLO model, which returns a list of bounding boxes identifying detected targets. These bounding boxes serve as inputs for SAM, which generates segmentation masks to outline the object pixels. Subsequently, these segmented object pixels are integrated into the base images based on the segmentation mask guidance.
Here, we elaborate on the detailed methodology.
\begin{itemize}
    \item \textbf{Object Detection with YOLO}: Object images are inputted into the YOLO model, specifically trained on Obj365, to detect objects belonging to predefined target categories fitting our scenario.
    \item \textbf{Segmentation with SAM}: Bounding boxes from YOLO are used as prompts for SAM to generate segmentation masks. These masks delineate object pixels, facilitating their extraction from the object images.
    \item \textbf{Integration with Base Images}: Extracted object pixels are seamlessly integrated into the corresponding regions of base images, aligning with the guidance provided by the segmentation masks.
\end{itemize}
During the SAM stage, while not every segmentation mask achieves perfection, each contributes to the overall objective of accurately segmenting the railway area rather than focusing on obstacles. To address challenges related to out-of-distribution (OOD) scenarios, we introduce random polygon generation with texture rendering from DTD. Additionally, object resizing and rescaling are applied to enrich image content and bolster model robustness.
The rescale follow the equation below: 
\begin{equation}
    h = \alpha\ast y+\beta
\end{equation}
\begin{equation}
    w = \frac{h}{H}\ast W
\end{equation}
where h,w and H,W are the shapes of the target obj and original obj, respectively. $\alpha$ and $\beta$ are hyper-parameters to adjust the scale. In our project, we set  $\alpha$ to 0.6 and $\beta$ to 30.
The value should be varied by the camera's position and it's parameters. The final generated samples are
demonstrated as \textbf{Fig .}\ref{sample}
% \begin{figure}[htb]
%     \centering
%     \includegraphics[width=0.45\textwidth]{figures/weather.png}
%     \caption{sample images of different weathers}
%     \label{fig:sample images}
% \end{figure}

% \begin{figure}[htb]
%     \centering
%     \includegraphics[width=0.45\textwidth]{figures/sample.png}
%     \caption{sample images of different weathers}
%     \label{fig:sample images}
% \end{figure}
\subsection{Optical-Flow}
Optical flow is based on the assumption that the intensity of a point in an image remains constant as it moves from one frame to the next. 
\begin{equation}
    I(x,y,t) = I(x+\Delta t,y+\Delta t, t+\Delta t) 
\end{equation}
% In our scenario, we use RAFT(Recurrent All-Pairs Field Transforms) as our preferred model, Which perform well in both tiny and large scale. In our scenario, the size of obstacles range from hundreds of pixels to less than 50. To use the RAFT model, two subsequent frames are needed. We generate two pseudo images $I_{t}$ and $I_{t+1}$ with same pasted target objects and a tiny position shift $\eta$.
In our scenario, we employ RAFT (Recurrent All-Pairs Field Transforms) as our chosen model, which demonstrates robust performance across a wide range of scales from tiny to large. The size of obstacles in our dataset varies, spanning from hundreds of pixels down to less than 50 pixels in size. Utilizing the RAFT model requires two consecutive frames for optical flow estimation. Accordingly, we generate two pseudo images $I_{t}$ and $I_{t+1}$, where the same target objects are pasted with a slight positional shift  $\eta$.
\begin{equation}
    Motion = \phi(I_{t},I_{t+1})
\end{equation}

\begin{equation}
    I_{t+1} = I_{t}(obj_{x}+\Delta x,obj_{y}+\Delta y) 
\end{equation}
We set $\Delta x$ and $\Delta y$ range between 5-10. The motion prediction will be leveraged as prior
information fused with pseudo image to train the model.

\section{Experiments}
\subsection{Dataset and Evaluation Metrics}
\noindent\textbf{Dataset} Our training dataset is consist of three parts: $obs\_{person}$, $obs\_{animals}$ and $obs\_{textures}$, namely person obstacles, animal obstacles and obstacles generated from texture polygons. The details are described as follow:
As for test dataset, we recollect images with various obstacles under different weather conditions in different distance to the camera.

\noindent\textbf{Metrics} $mIoU$ is used to evaluate the performance of our model. $mIoU$ refers to the Mean Intersection over union, which is a widely used metric in segmentation task. It can be calculated as follow:
\begin{equation}
    IoU_{i} = \frac{TP_{i}}{TP_{i}+FP_{i}+FN_{i}}
\end{equation}
\begin{equation}
    mIoU = \frac{1}{n}\sum_{i=0}^{n} IoU_{i}
\end{equation}
pixel accuracy is also a metric to evaluate the segmentation models.
\begin{equation}
    Pixel\_{accuracy} = \frac{N\_{corr}}{N\_{total}}
\end{equation}
where $N\_{corr}$ is the number of correctly classified pixels, $N\_{total}$ is the number of total pixels.

\vspace{1cm}
\renewcommand{\arraystretch}{1.5} 
\begin{table}
    \centering
    \begin{tabular}{c|c|c|c}
      \toprule
      Name & Volume & Dis(m)&  Category \\
      \hline
      $obs\_{person}$ & 4000 & 0-70&person\\
      \hline
      $obs\_{animal}$ & 4000 &0-70&cow,horse,deer\\
      \hline
      $obs\_{texture}$  & 2000&  0-70 &see DTD\\
      \hline
      $val\_{near}$  & 200&  0-20 &person,rock,board\\
      \hline
      $val\_{middle}$  & 200& 20-50 &person,rock,board\\
      \hline
      $val\_{far}$  & 200&  50-70 &person,rock,board\\
      \bottomrule
    
    \end{tabular}
    \caption{Datasets description.}
    \label{data}
\end{table}

\subsection{Implementation Details}
Our method is implemented using the PyTorch framework and the model is trained on an RTX 4070Ti. We select Jaccard loss as the loss function and AdamW as the optimizer. The batch size is set to 8 and the number of epochs to 20. Data transformations include horizontal flip, coarse dropout, and random brightness contrast adjustments.

\subsection{Results}
% To validate the performance of our approach, we conduct experiment on our our three self-collected dataset:  
% val\_near, val\_mid, val\_far. The details are described in \textbf{Table}\ref{tab:my_label}. Basic training dataset contain 10000 images(4000+4000+2000), to fully validate the effect of the size of generated images, we add 10\% and 50\% of basic image size in row 4 and 5. The results show that both raft and pure segmentation based approach have the ability to segment obstacles in our experiment railway area. Fusing the raft and pseudo images can boost the performance of the model. With more generated images added to training dataset, the model gradually reaches its limit.
To validate the performance of our approach, we conduct experiments on our three self-collected datasets: val\_near, val\_mid, and val\_far. The details are described in \textbf{Table} \ref{data}. The basic training dataset contains 10,000 images (4,000+4,000+2,000). To fully assess the impact of the number of generated images, we increase the dataset size by 10\% and 50\% in rows 4 and 5. 

The results are illustrated in \textbf{Table}\ref{results}, which show that both RAFT and segmentation-based approaches can effectively segment obstacles in our railway area experiments. Combining RAFT and pseudo-images enhances model performance. As more generated images are added to the training dataset, the model's performance gradually reaches its limit.
\begin{table}[htb]
    \centering
    \begin{tabular}{ccccc}
    \toprule
         &\textbf{val\_near}&\textbf{val\_mid}&\textbf{val\_far}  \\
    \midrule
    \midrule
      yolov5   &0.744&0.623&0.457\\
      \hline
       Raft   &0.717&0.636&0.585  \\
    \hline
      DeepLab   &0.825&0.817&0.747  \\
    \hline
      DeepLab+Raft  &0.843&0.828&0.709  \\
    \hline
     DeepLab+Raft+10\%   &0.837&0.843 & 0.724 \\ 
     \hline
      DeepLab+Raft+50\%   &\textbf{0.863}&\textbf{0.851} &\textbf{0.802} \\ 
    \bottomrule
    \end{tabular}
    \caption{Experiments results.}
    \label{results}
\end{table}
\subsection{Ablation Study}
 We conduct ablation experiment to validate the effect of different target objects. The results are demonstrated as \textbf{Table} \ref{ablation} Comparing the row 1,2,3 with row 4, we can find that each obs dataset contributes to improving the robustness and accuracy of the model.
\begin{table}[htb]
    \centering
    \begin{tabular}{c|cccc}
    \toprule
         &obs\_person&obs\_animal&obs\_texture& mIoU \\
    \midrule
        1 &\ding{51} & \ding{51}& \ding{55}&0.781\\
        2 &\ding{51} &\ding{55}&\ding{51} &0.817\\
        3 &\ding{55} &\ding{51} & \ding{51}&0.732\\
        4 &\ding{51} &\ding{51} &\ding{51} &0.849\\
    \bottomrule
    \end{tabular}
    \caption{Ablation study.}
    \label{ablation}
\end{table}

\section{Conclusion}
This paper introduces a universal segmentation model based on a semi-supervised approach. To address out-of-distribution (OOD) challenges, we generate highly realistic pseudo images instead of relying on manual pixel-level annotations. Additionally, we enhance performance by incorporating optical flow techniques. Experimental results demonstrate satisfactory performance across various potential objects.

\bibliographystyle{plain}

\bibliography{ref}

\begin{thebibliography}{10}

\bibitem{boussik2021railway}
Amine Boussik, Wael Ben-Messaoud, Smail Niar, and Abdelmalik Taleb-Ahmed.
\newblock Railway obstacle detection using unsupervised learning: An exploratory study.
\newblock In {\em 2021 IEEE Intelligent Vehicles Symposium (IV)}, pages 660--667. IEEE, 2021.

\bibitem{brucker2023local}
Matthias Brucker, Andrei Cramariuc, Cornelius Von~Einem, Roland Siegwart, and Cesar Cadena.
\newblock Local and global information in obstacle detection on railway tracks.
\newblock In {\em 2023 IEEE/RSJ International Conference on Intelligent Robots and Systems (IROS)}, pages 9049--9056. IEEE, 2023.

\bibitem{cai2022enable}
Han Cai, Ji~Lin, Yujun Lin, Zhijian Liu, Haotian Tang, Hanrui Wang, Ligeng Zhu, and Song Han.
\newblock Enable deep learning on mobile devices: Methods, systems, and applications.
\newblock {\em ACM Transactions on Design Automation of Electronic Systems (TODAES)}, 27(3):1--50, 2022.

\bibitem{cimpoi2014describing}
Mircea Cimpoi, Subhransu Maji, Iasonas Kokkinos, Sammy Mohamed, and Andrea Vedaldi.
\newblock Describing textures in the wild.
\newblock In {\em Proceedings of the IEEE conference on computer vision and pattern recognition}, pages 3606--3613, 2014.

\bibitem{fedynyak2024devos}
Volodymyr Fedynyak, Yaroslav Romanus, Bohdan Hlovatskyi, Bohdan Sydor, Oles Dobosevych, Igor Babin, and Roman Riazantsev.
\newblock Devos: Flow-guided deformable transformer for video object segmentation.
\newblock In {\em Proceedings of the IEEE/CVF Winter Conference on Applications of Computer Vision}, pages 240--249, 2024.

\bibitem{guo2023enhancing}
Qiushi Guo, Yifan Chen, and Shisha Liao.
\newblock Enhancing mobile privacy and security: A face skin patch-based anti-spoofing approach.
\newblock In {\em 2023 IEEE International Conference on Cloud Computing Technology and Science (CloudCom)}, pages 52--57. IEEE, 2023.

\bibitem{kirillov2023segment}
Alexander Kirillov, Eric Mintun, Nikhila Ravi, Hanzi Mao, Chloe Rolland, Laura Gustafson, Tete Xiao, Spencer Whitehead, Alexander~C Berg, Wan-Yen Lo, et~al.
\newblock Segment anything.
\newblock In {\em Proceedings of the IEEE/CVF International Conference on Computer Vision}, pages 4015--4026, 2023.

\bibitem{redmon2016you}
Joseph Redmon, Santosh Divvala, Ross Girshick, and Ali Farhadi.
\newblock You only look once: Unified, real-time object detection.
\newblock In {\em Proceedings of the IEEE conference on computer vision and pattern recognition}, pages 779--788, 2016.

\bibitem{flood2019detecting}
Cem Sazara, Mecit Cetin, and Khan~M Iftekharuddin.
\newblock Detecting floodwater on roadways from image data with handcrafted features and deep transfer learning.
\newblock In {\em 2019 IEEE intelligent transportation systems conference (ITSC)}, pages 804--809. IEEE, 2019.

\bibitem{sevilla2016optical}
Laura Sevilla-Lara, Deqing Sun, Varun Jampani, and Michael~J Black.
\newblock Optical flow with semantic segmentation and localized layers.
\newblock In {\em Proceedings of the IEEE conference on computer vision and pattern recognition}, pages 3889--3898, 2016.

\bibitem{shao2019objects365}
Shuai Shao, Zeming Li, Tianyuan Zhang, Chao Peng, Gang Yu, Xiangyu Zhang, Jing Li, and Jian Sun.
\newblock Objects365: A large-scale, high-quality dataset for object detection.
\newblock In {\em Proceedings of the IEEE/CVF international conference on computer vision}, pages 8430--8439, 2019.

\bibitem{zhang2023automatic}
Qiang Zhang, Fei Yan, Weina Song, Rui Wang, and Gen Li.
\newblock Automatic obstacle detection method for the train based on deep learning.
\newblock {\em Sustainability}, 15(2):1184, 2023.

\end{thebibliography}
\end{document}